\documentclass[10pt,twocolumn,letterpaper]{article}
\usepackage[pagenumbers]{cvpr}

\usepackage{graphicx}
\usepackage{amsmath,amssymb,amsthm}
\usepackage{booktabs,multirow,makecell}
\usepackage{xcolor}
\usepackage{colortbl}
\makeatletter
\newcommand{\trimrowfill}{%
  \gdef\CT@cell@color{\@tempdimc\z@\global\let\CT@cell@color\relax}}
\makeatother
\definecolor{cvprblue}{rgb}{0.21,0.49,0.74}
\definecolor{stdred}{HTML}{B64342}
\definecolor{presblue}{HTML}{0072B2}
\usepackage[pagebackref,breaklinks,colorlinks,allcolors=cvprblue]{hyperref}
\usepackage[capitalize]{cleveref}

\newcommand{\Nec}{\mathcal{N}}
\newcommand{\Suf}{\Sigma}

\newcommand{\Dfut}{D_{\mathrm{fut}}}

\newcommand{\nbf}[1]{\par\textbf{#1}\hspace{0.5em}}

\title{SteerTrace: Tracing How Target Attacks Steer World Action Models}

\author{
Yukun Dai$^{1}$ \quad Tianshi Wang$^{1}$ \quad Fengling Li$^{2}$ \quad Jingjing Li$^{3}$ \quad Xin Li$^{4}$ \quad Lei Zhu$^{1}$\\[4pt]
$^{1}$Tongji University \quad
$^{2}$Mohamed bin Zayed University of Artificial Intelligence\\
$^{3}$University of Electronic Science and Technology of China \quad
$^{4}$Shanghai Artificial Intelligence Laboratory
}

\begin{document}
\maketitle

\begin{abstract}
World action models (WAMs) predict future videos and actions with one backbone, and their imagined future is often seen as a safety signal. We introduce SteerTrace, a targeted universal attack on WAMs that also traces its effect inside the model. One $\ell_\infty$-bounded ($4/255$) perturbation per model, shared across frames, tasks, and camera views, pushes the executed action toward an attacker-chosen target. On eight WAMs from four families (FastWAM, Faster-WAM, OpenWAM and LingBot-VA) on LIBERO and RoboTwin, it drives task success close to zero and moves the attacked action toward the target. Same-budget noise and BadWAM produce little action shift in the target direction. We use the signed targeted effect to measure how much of the effect each intervention removes or reproduces. Under the standard attack, restoring future tokens removes $61$--$94\%$ of the targeted effect in the five models whose action experts attend to them. When the attack is constrained to keep the predicted future close to the clean prediction, the targeted effect can become more dependent on current-frame tokens. In three of these five models, restoring current-frame tokens removes $87$--$99\%$ of the effect, compared with $9$--$15\%$ for future tokens. In six models, restoring just $0.36\%$ of the FFN neurons in layers 10--29 of the video expert removes $72$--$96\%$ of the targeted effect. Residual directions 2--16 hold only $11$--$18\%$ of the shift energy, yet removing them cancels most of the targeted effect.
\end{abstract}

\section{Introduction}
\label{sec:intro}
\begin{figure}[t]
\centering
\includegraphics[width=\columnwidth]{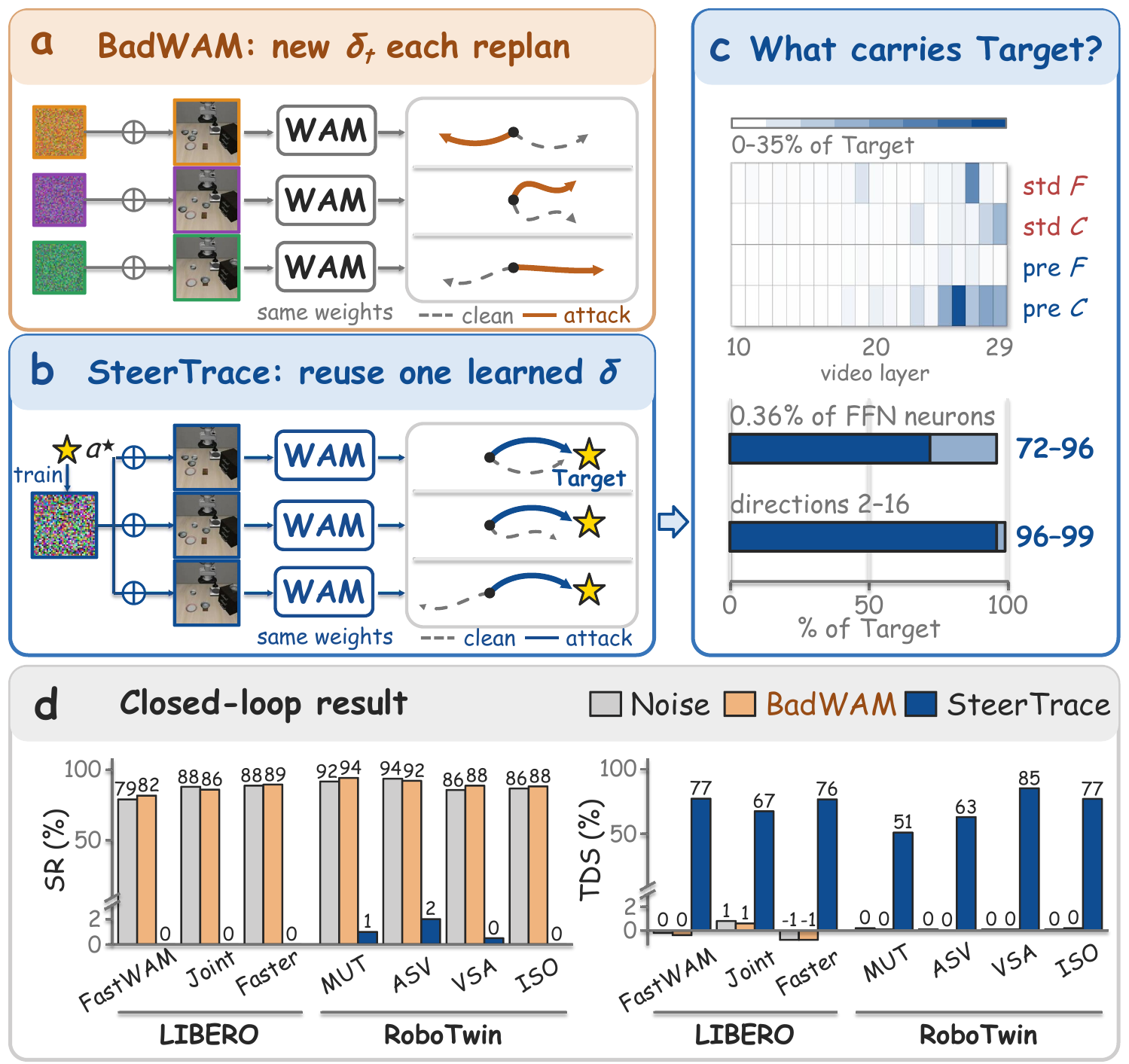}
\caption{\textbf{Targeted universal attacks that reveal how world action models are steered.} (a) Prior attacks~\cite{badwam} optimize a perturbation for every input and maximize the distance between each action (solid) and its clean value (dashed), with no control over where the action goes. (b) One perturbation pushes every output action toward an attacker-chosen target action $a^\star$, yielding a targeted effect (Target, $T$). (c) Using this targeted effect as the measure: share of $T$ removed by restoring one video layer of the future ($\mathcal{F}$) or current-frame ($\mathcal{C}$) tokens in FastWAM-Joint, a few FFN neurons, or residual directions 2--16. (d) Closed-loop success rate (SR) and shift toward the target (TDS). Value axes are broken at 3.}
\label{fig:teaser}
\end{figure}

World action models (WAMs)~\cite{fastwam,fasterwam,openwam,lingbotva} couple a video generator with an action decoder. Unlike a vision-language-action model (VLA), which maps observations to actions directly, a WAM is trained to predict how the scene evolves, and most WAMs also \emph{imagine} the next few frames at test time and condition the action chunk on that imagination through attention. The imagined future is often argued to make policies more grounded, and it brings an apparent safety benefit: a human or an automated runtime monitor can inspect it before each chunk is executed and reject the chunk if the future looks wrong~\cite{trustedimagination,badwam}.

Two existing attacks study how the imagined future relates to the executed action. BadWAM~\cite{badwam} gives a first indication that a WAM can ``dream right but act wrong'': small perturbations make the task fail while the predicted future stays close to the clean prediction. Attacking the Trusted Imagination~\cite{trustedimagination} shows that the reverse also holds: perturbations corrupt the imagined future while the action of the model barely changes. Both attacks optimize a new perturbation for every input, and neither can choose what the WAM does. It therefore remains open whether one perturbation can steer a WAM while its imagined future looks clean, and what role the imagined future plays in carrying such an attack to the action.

We study a different threat: one fixed perturbation per model that steers the executed action. A single $\ell_\infty$-bounded ($\le 4/255$) image perturbation, shared across every frame, task and camera, pushes the end effector of the executed action toward an attacker-chosen target. It is trained in a white-box setting and needs no further optimization at deployment. Like BadWAM, the attack has a variant that also keeps the predicted future close to clean. Unlike BadWAM, this variant is universal and targeted. Because the attack has a target, its effect is a signed quantity, and every intervention inside the model has a quantitative outcome: the fraction of the targeted effect it removes or reproduces. We use this to trace the attack at three levels: the tokens whose keys and values the action expert reads, the FFN neurons of the video expert, and the directions of the residual stream. We call the attack together with this trace SteerTrace.

We attack eight WAMs from four families on LIBERO~\cite{libero} and RoboTwin~\cite{robotwin2}: FastWAM~\cite{fastwam} and its joint variant FastWAM-Joint, Faster-WAM~\cite{fasterwam}, OpenWAM in four variants that differ only in the attention mask~\cite{openwam}, and the autoregressive LingBot-VA~\cite{lingbotva}. 

\begin{itemize}
\item \textbf{A universal perturbation that steers the executed action.} To our knowledge, this is the first universal perturbation that moves the executed action of a WAM toward an attacker-chosen target. It drives task success close to zero across the eight WAMs we evaluate (\cref{sec:main}). 
\item \textbf{The imagined future is the attack's preferred pathway.} In all five models whose action expert attends to the imagined future, unconstrained attacks pass through it: restoring the future tokens removes $61$--$94\%$ of the targeted effect. When the predicted future is preserved near clean, the attack moves to the current frame in FastWAM-Joint, Faster-WAM and OpenWAM-MUT (\cref{fig:teaser,sec:route}).
\item \textbf{A small set of neurons carries the targeted effect.} $0.36\%$ of the FFN neurons carry $72$--$96\%$ of the targeted effect in six models, and the set changes when the attack changes pathway (\cref{sec:neurons}).
\item \textbf{Low-energy directions carry the targeted effect.} Directions 2--16 of the residual shift hold only $11$--$18\%$ of its energy, yet removing them cancels $96$--$99\%$ of the targeted effect in seven models (\cref{sec:directions}).
\end{itemize}

\section{Related Work}
\label{sec:related}

\nbf{World action models.}
Video generation has served as a policy interface since UniPi~\cite{du2023unipi}, video pre-training improves manipulation policies~\cite{wu2024gr1,hu2025vpp}, and large video models~\cite{bruce2024genie,agarwal2025cosmos,wan} now serve as backbones of WAMs that predict future frames and actions together~\cite{zhu2025uwm,bi2025motus,kim2026cosmospolicy,ye2026dreamzero,gigaworld2026}. These WAMs differ in how the action reads the imagined future: it attends to the imagined frames in the joint FastWAM~\cite{fastwam} and in the autoregressive LingBot-VA~\cite{lingbotva}, to a few selected future representations in Faster-WAM~\cite{fasterwam}, and not at all in the video-free FastWAM, which uses future prediction only as a training objective. OpenWAM~\cite{openwam} compares WAM designs in controlled experiments, including how the action relates to the imagined future. Our attacks span this range.

\nbf{Adversarial attacks on WAMs.}
Adversarial examples~\cite{szegedy2014intriguing,goodfellow2015explaining,madry2018pgd} extend to universal perturbations~\cite{moosavi2017uap,doubletargeteduap}, physical patches~\cite{brown2017patch,eykholt2018robust} and sequential decision making~\cite{huang2017rl,gleave2020adversarial}, and on vision-language-action models they cause task failure~\cite{cheng2024manipulation,lu2026obeypatch}, targeted motion~\cite{wang2024vlaattack,huang2026trap} or backdoor behavior~\cite{zhou2025badvla}. WAMs add an attack surface that these policies lack: the imagined future. BadWAM~\cite{badwam} keeps this future close to clean while making the task fail, and Attacking the Trusted Imagination~\cite{trustedimagination} corrupts it and detects such corruption with a denoiser. Both optimize a perturbation for every input. TAPDreamer~\cite{tapdreamer} and PASA-KV~\cite{pasakv} instead attack how the current observation is encoded, with patches that change the representation of a public visual encoder and the visual keys and values attended to by the action expert, respectively. A detailed comparison with these two concurrent attacks is in the Supplementary. None of these attacks steers the executed action toward a chosen target. Ours is, to our knowledge, the first targeted universal attack on WAMs, and we use it to examine what role the imagined future plays in carrying an attack to the action.

\nbf{Mechanistic analysis.}
Causal interventions on hidden states, including activation patching and causal tracing~\cite{vig2020causal,meng2022rome}, path patching~\cite{wang2023ioi} and automated circuit discovery~\cite{conmy2023acdc}, locate the components that drive a model output, and attribution patching approximates them to first order~\cite{nanda2022attribution,syed2023attribution,kramar2024atp}. At the level of units, FFN layers act as key--value memories~\cite{geva2021ffn}, and individual neurons carry specific knowledge~\cite{dai2022knowledge} or safety behavior~\cite{chen2024safetyneurons}. At the level of directions, behaviors such as refusal are mediated by low-dimensional subspaces of the residual stream~\cite{arditi2024refusal,zou2023repe,turner2023activation}, and features need not align with individual neurons~\cite{elhage2022superposition}. Adversarial perturbations exploit non-robust input features~\cite{ilyas2019features}. These analyses have not been used to trace a targeted effect on the action of a WAM.

\section{Method}
\label{sec:method}

\begin{figure*}[t]
\centering
\includegraphics[width=\textwidth]{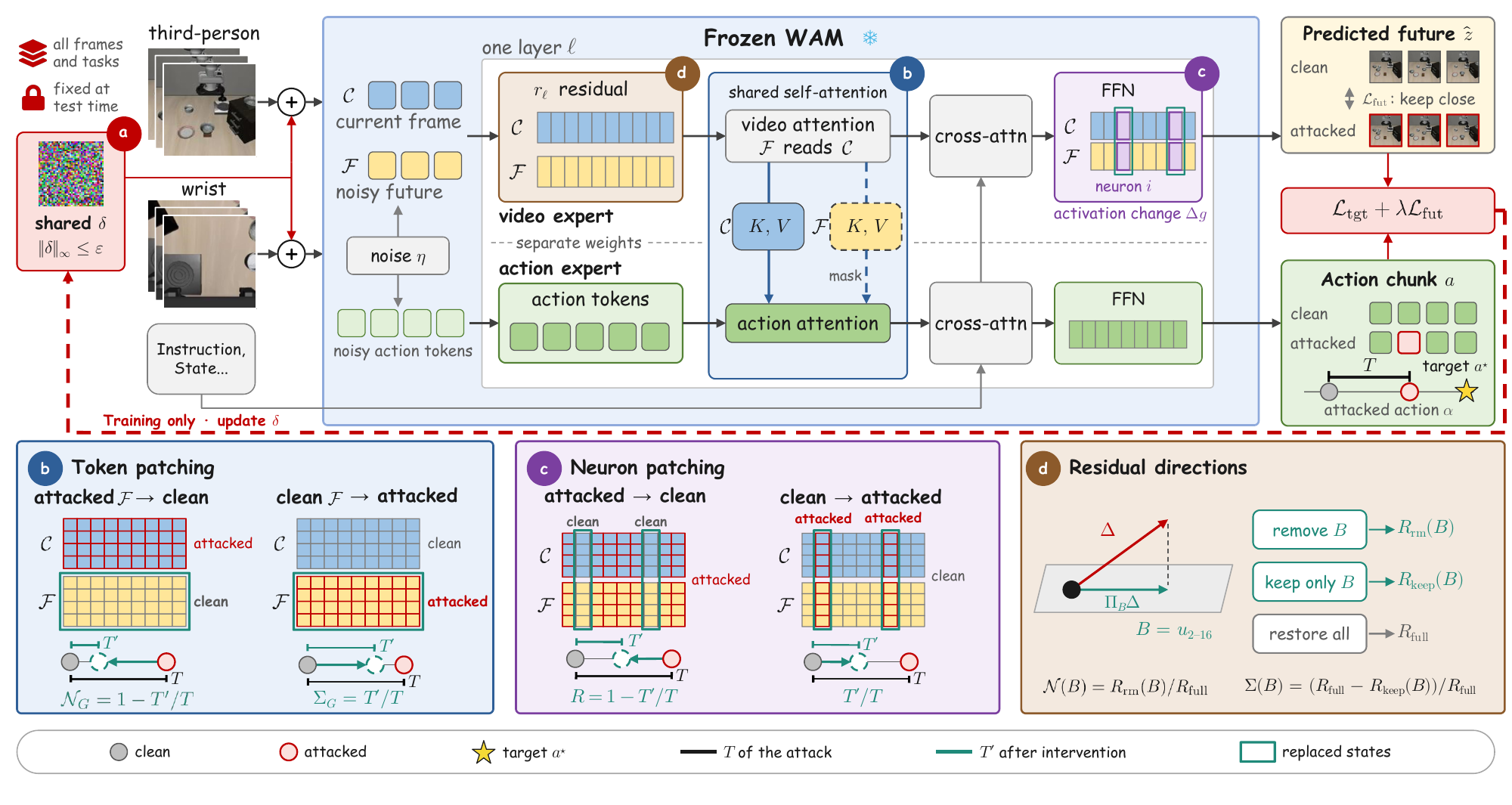}
\caption{\textbf{Framework overview.} \textbf{Top:} (a)~one perturbation $\delta$ is added to every camera frame and is the only quantity optimized. The WAM is frozen. The action tokens read the K/V of the current-frame tokens $\mathcal{C}$ and, if the mask allows (dashed), of the future tokens $\mathcal{F}$. \textbf{Bottom:} three analyses intervene at the sites b--d and measure the remaining targeted effect $T'$. (b)~For one token group, we restore clean K/V in the attacked run or inject attacked K/V into the clean run. (c)~For the top-$k$ FFN neurons ranked by SteerTrace, we restore clean activations in the attacked run or inject attacked activations into the clean run. (d)~We remove or keep a band $B$ of the residual shift $\Delta$. State outlines and action-value circles are gray for clean and red for attacked.
Green outlines mark replaced states.}
\label{fig:framework}
\end{figure*}

We first describe the models we attack and the threat model (\cref{sec:prelim}), then the targeted universal perturbation (\cref{sec:attack}) and the mechanistic analyses we apply to it (\cref{sec:mech}). 
 
\subsection{Preliminaries and threat model}
\label{sec:prelim}
\nbf{World action models.}
Given camera observations $x$ and an instruction $L$, a WAM $\pi_\theta$ predicts future video latents $\hat z$, which we call the imagined future, together with an action chunk $a=(a_1,\dots,a_H)$ of $H$ actions $a_h\in\mathbb{R}^{m}$. The attack acts on one of the $m$ action outputs, the attacked action, which we write $\alpha_h\in\mathbb{R}$,
\begin{equation}
(\hat z, a)=\pi_\theta(x, L).
\end{equation}
The WAMs we study pair a video expert with an action expert in a Mixture-of-Transformers~\cite{fastwam,fasterwam,lingbotva} (\cref{fig:framework}). Each expert has its own weights, including its own FFN, and the two share self-attention. The observation is encoded into current-frame tokens $\mathcal{C}$, and the imagined future is represented by future tokens $\mathcal{F}$. The action tokens start from noise and are denoised by the action expert over several sampling steps. In the shared attention, they read the keys and values (K/V) of $\mathcal{C}$ and, if the attention mask allows, of $\mathcal{F}$. A perturbation of the input therefore reaches the action either through $\mathcal{C}$ or through $\mathcal{F}$, which itself attends to $\mathcal{C}$.

\nbf{Threat model.}
We attack a frozen WAM $\pi_\theta$.
\emph{Goal.} The attacker wants to move the attacked action toward a chosen target action $a^\star$.
\emph{Access.} The attacker has white-box access to $\pi_\theta$, including its parameters and gradients, and uses frames from its training data to optimize the perturbation.
\emph{Scope.} One fixed perturbation $\delta$ with $\|\delta\|_\infty\le\epsilon$ is added to every frame of every camera view in every task of a suite $\mathcal{S}=\{\tau_1,\tau_2,\dots\}$. The attacker does not change the weights, the instruction, or the scene. Unlike per-input attacks~\cite{badwam}, $\delta$ is neither optimized for individual inputs nor updated during deployment.
 
\subsection{Targeted universal perturbation}
\label{sec:attack}
 
\nbf{Problem formulation.}
Let $\delta$ denote a universal perturbation shared by all camera views, so that the attacked observation is $x+\delta$. Each training frame supplies a pair $\xi=(x,L)$ of an observation and its instruction. We optimize a single $\delta$ over the task distribution,
\begin{equation}
\delta^{\star}=\mathop{\mathrm{argmin}}_{\|\delta\|_\infty\le\epsilon}\;
\mathbb{E}_{\xi\sim\mathcal{D}}
\Big[\mathcal{L}_{\mathrm{tgt}}(\delta;\xi,a^{\star})+\lambda\,\mathcal{L}_{\mathrm{fut}}(\delta;\xi)\Big],
\label{eq:std}
\end{equation}
where $\mathcal{D}$ is the distribution of training frames over the tasks in $\mathcal{S}$, $a^\star$ is the chosen target action, $\mathcal{L}_{\mathrm{tgt}}$ is the target loss, $\mathcal{L}_{\mathrm{fut}}$ is the future-preserving loss and $\lambda\ge 0$ is its weight. We call the attack \emph{standard} when $\lambda=0$ and \emph{future-preserving} (\emph{preserve} for short) when $\lambda>0$.
 
\nbf{Target loss.}
In all WAMs we attack, the action expert is trained by flow matching and generates the chunk with an Euler sampler. Sampling starts from a Gaussian noise chunk $a^{(0)}=\eta\sim\mathcal{N}(0,I)$, $\eta\in\mathbb{R}^{H\times m}$, and integrates the predicted velocity with $S$ Euler steps on the model's own flow-time grid $1=t_0>t_1>\dots>t_S=0$,
\begin{equation}
a^{(s+1)}=a^{(s)}+(t_{s+1}-t_s)\,v_\theta\big(a^{(s)},t_s;x+\delta,L\big),
\label{eq:sampler}
\end{equation}
where $s=0,\dots,S-1$ and $v_\theta$ is the velocity field of the frozen action expert, which sees the attacked observation only through the shared attention. The final iterate $a(x+\delta,L;\eta)=a^{(S)}$ is the predicted chunk. Unlike flow-matching training, which regresses $v_\theta$ at a single noisy point, we place the loss on the sampled chunk itself, so that it measures the executed action and its gradient passes through every denoising step. The target loss is the squared error between $a^\star$ and the attacked action of this chunk, in normalized action space,
\begin{equation}
\mathcal{L}_{\mathrm{tgt}}(\delta;\xi,a^{\star})=\mathbb{E}_{\eta}\Big[\frac{1}{H}\sum_{h=1}^{H}\big(\alpha_h(x+\delta,L;\eta)-a^{\star}\big)^2\Big],
\label{eq:tgt}
\end{equation}
where $h$ indexes the action steps. All other action outputs are excluded from the loss and are not constrained.
 
\nbf{Future-preserving loss.}
The future-preserving loss penalizes the deviation of the imagined future from the clean prediction for the same frame,
\begin{equation}
\mathcal{L}_{\mathrm{fut}}(\delta;\xi)=\big\|\hat z(x+\delta,L)-\mathrm{sg}\big[\hat z(x,L)\big]\big\|_2^2,
\label{eq:preserve}
\end{equation}
where $\mathrm{sg}$ is the stop-gradient operator.
 
\nbf{Optimization.}
We compute both losses with the full sampler of the model and back-propagate through all of its denoising steps. At every iteration we draw new sampling noise, shared by the clean and the attacked prediction, and after each update we clip $\delta$ to $[-\epsilon,\epsilon]$.
 
\subsection{Mechanistic analysis}
\label{sec:mech}
Because the attack has a signed target, every intervention has a quantitative outcome, the fraction of the targeted effect it removes or reproduces. We measure it for the token groups the action expert attends to, the FFN neurons of the video expert and the directions of its residual stream.

\nbf{Targeted effect as the measure.} Each analysis pairs an attacked run on $(x+\delta,L)$ with a clean run on $(x,L)$ under the same sampling noise, and an intervention $\iota$ replaces internal states of one run with those of the other. For a run on $(\tilde x,L)$ with intervention $\iota$ and attacked actions $\tilde\alpha_h$, the targeted effect is the part of the gap between the clean prediction and $a^\star$ that the run closes,
\begin{equation}
T(\tilde x;\iota)=\frac{1}{H}\sum_{h=1}^{H}\Big(\big|a^{\star}-\alpha_h(x,L)\big|-\big|a^{\star}-\tilde\alpha_h\big|\Big).
\label{eq:T}
\end{equation}
We write $T=T(x+\delta;\varnothing)$ for the attacked run and $T'$ for a run with an intervention. An intervention that restores clean states removes the fraction
\begin{equation}
R=1-T'/T.
\label{eq:R}
\end{equation}

\nbf{Token-group patching.}
We split the video tokens into the current-frame tokens $\mathcal{C}$ and the future tokens $\mathcal{F}$. For a group $G$, we replace the K/V that the action tokens attend to, at every layer and denoising step or in a range of layers. Restoring the clean K/V in the attacked run tests necessity, and injecting the attacked K/V into the clean run tests sufficiency,
\begin{align}
\mathcal{N}_G&=1-\frac{T\big(x{+}\delta;\ \mathrm{KV}_G\!\leftarrow\!\mathrm{KV}_G^{c}\big)}{T},\label{eq:nec}\\
\Sigma_G&=\frac{T\big(x;\ \mathrm{KV}_G\!\leftarrow\!\mathrm{KV}_G^{\mathrm{adv}}\big)}{T}.\label{eq:suf}
\end{align}

\nbf{Neuron attribution and patching.}
Token groups locate \emph{where} the attack enters the action expert, and neurons locate \emph{what} it changes in the video expert. A neuron is one entry $i$ of the FFN hidden activation $g_\ell=\phi(W_1 y)$ of video layer $\ell$. It takes a value $g_{\ell,i}^{(s,n)}$ at every token $n$ and forward pass $s$ of the video expert, and we attribute and patch these values jointly. Inspired by attribution patching~\cite{nanda2022attribution,kramar2024atp}, we score each neuron by a first-order estimate of how much $T$ falls when it is restored to clean. As the neuron lies in the video expert and $T$ is computed from the output of the action expert, we take the gradient through the shared attention and every denoising step,
\begin{equation}
A_{\ell,i}=\mathbb{E}_{x\in\mathcal{X}_{\mathrm{sel}}}\Big[\sum_{s,n}\Delta g_{\ell,i}^{(s,n)}\,\frac{\partial T}{\partial g_{\ell,i}^{(s,n)}}\Big],
\label{eq:attr}
\end{equation}
where $\Delta g=g(x+\delta,L)-g(x,L)$ is the \emph{activation change}, the gradient is evaluated in the attacked run, and $\mathcal{X}_{\mathrm{sel}}$ is a selection set of frames pooled over tasks. We keep the top $k$ neurons of a set of video layers $\Lambda$ as the set $\mathcal{U}$ and, on disjoint test frames, rerun the attacked input with
\begin{equation}
g_{\ell,i}^{(s,n)}(x+\delta,L)\leftarrow g_{\ell,i}^{(s,n)}(x,L)
\label{eq:npatch}
\end{equation}
for all $(\ell,i)\in\mathcal{U}$ and all $s,n$, and report $R$. We compare with the common rule of taking the $k$ neurons with the largest $|\Delta g|$~\cite{chen2024safetyneurons} and with $k$ random neurons, and test sharing by selecting $\mathcal{U}$ without the evaluated task.

\nbf{Residual-subspace ablation.}
Let $\Delta_{\ell,n}=r_{\ell,n}(x+\delta,L)-r_{\ell,n}(x,L)$ be the \emph{residual shift}, the change of the residual-stream input of video layer $\ell\in\Lambda$ at token $n$, and $\|\Delta_{\ell,n}\|^2$ its energy. Whereas the activation change $\Delta g$ concerns a single FFN neuron, the residual shift is the change of the whole residual-stream vector passed between layers. The eigenvectors $u_1,u_2,\dots$ of $M_\ell=\mathbb{E}_{\tau,x,n}[\Delta_{\ell,n}\Delta_{\ell,n}^{\top}]$, ordered by decreasing eigenvalue, are the directions with the most shift energy, and we analyze bands $B$ of consecutive rank, such as $u_{2\text{--}16}$. With $\Pi_B$ the orthogonal projection onto $B$, $\rho(B)=\mathbb{E}\|\Pi_B\Delta_{\ell,n}\|^2/\mathbb{E}\|\Delta_{\ell,n}\|^2$ is the energy share of $B$. At every layer in $\Lambda$, we replace the attacked residual $r$ by $r-\Pi_B(r-r^{c})$ (band removed), $r^{c}+\Pi_B(r-r^{c})$ (only the band kept) or the clean residual $r^{c}$, which gives $R_{\mathrm{rm}}(B)$, $R_{\mathrm{keep}}(B)$ and $R_{\mathrm{full}}$. The band analogues of $\mathcal{N}_G$ and $\Sigma_G$ are
\begin{align}
\mathcal{N}(B)&=\frac{R_{\mathrm{rm}}(B)}{R_{\mathrm{full}}},\\
\Sigma(B)&=\frac{R_{\mathrm{full}}-R_{\mathrm{keep}}(B)}{R_{\mathrm{full}}}.
\end{align}
We compare with random subspaces of equal dimension, and test sharing by estimating $M_\ell$ without the evaluated task.

\section{Experimental Setup}
\label{sec:setup}

\nbf{Victim models.}
We attack eight WAMs from four families. These are FastWAM~\cite{fastwam}, whose action expert does not attend to the imagined future, and its joint variant FastWAM-Joint. We further attack Faster-WAM~\cite{fasterwam}, the autoregressive LingBot-VA~\cite{lingbotva}, and four OpenWAM~\cite{openwam} variants, \texttt{mutual}, \texttt{action\_sees\_video}, \texttt{video\_sees\_action} and \texttt{isolated}. The five models whose action expert attends to the imagined future are FastWAM-Joint, Faster-WAM, LingBot-VA, \texttt{mutual} and \texttt{action\_sees\_video}.

\nbf{Benchmarks.}
We evaluate FastWAM, Faster-WAM and LingBot-VA on LIBERO~\cite{libero}, which has four suites (Spatial, Object, Goal and Long) of 10 tasks each. We evaluate FastWAM, Faster-WAM, and the four OpenWAM variants on the 50 tasks of RoboTwin~\cite{robotwin2}. 

\nbf{Attack settings.}
We set the target action $a^\star$ to the largest value of the attacked action in the dataset. The budget is $\epsilon=4/255$ in pixel scale, and one $\delta$ is shared by all camera views, two on LIBERO and three on RoboTwin. The preserve attack uses $\lambda=1$ on LIBERO and $\lambda=0.1$ on RoboTwin.

\nbf{Metrics.}
We report the task success rate (SR) and the target direction shift (TDS). TDS measures how far the attack moves the executed action toward $a^\star$: at every executed step, we compare the attacked action under $\delta$ with that of a paired query without $\delta$ on the same observation, and average over steps and episodes,
\begin{equation}
\mathrm{TDS}=\frac{\mathbb{E}\big[\alpha(x+\delta,L)-\alpha(x,L)\big]}{r}\times 100\%,
\label{eq:tds}
\end{equation}
where $\alpha$ is the executed attacked action and $r$ is half its range in the normalized action space. TDS is positive when the action moves toward $a^\star$ and stays near zero for an attack without a target direction. For preserve, we report the targeted effect $T$ of \cref{eq:T} and the relative deviation of the imagined future,
\begin{equation}
\Dfut=\frac{\|\hat z(x+\delta,L)-\hat z(x,L)\|_2}{\|\hat z(x,L)\|_2}.
\label{eq:dfut}
\end{equation}
We also compare the decoded future frames with the clean prediction using LPIPS~\cite{zhang2018lpips}, a learned perceptual distance.

\section{Results}
\label{sec:results}

\subsection{Targeted attack effectiveness}
\label{sec:main}

We evaluate whether the perturbation both disrupts task execution and steers actions toward the target.

\begin{table}[t]
\caption{\textbf{Attack effectiveness on LIBERO.} SR is the task success rate (\%) on each suite. TDS is the shift of the executed attacked action toward $a^\star$ relative to a paired clean query, in per cent of the half range of the normalized action. Clean TDS compares paired unperturbed rollouts that differ only in the sampling seed. For perturbed conditions, $\pm$ is the error across repeated runs. Noise is same-budget Gaussian noise clipped to $\|\delta\|_\infty\le4/255$. BadWAM is the attack of~\cite{badwam}, restricted to the same budget $\epsilon$ as ours. Since it has no target direction, its TDS is measured along the same axis. Standard ($\lambda=0$) and preserve ($\lambda=1$) are the two variants of SteerTrace, with $\epsilon=4/255$.}
\label{tab:main}
\centering
\footnotesize
\setlength{\tabcolsep}{3pt}
\resizebox{\columnwidth}{!}{%
\begin{tabular}{@{}ll*{4}{cr@{}l}@{}}
\toprule
 & & \multicolumn{3}{c}{Spatial} & \multicolumn{3}{c}{Object} & \multicolumn{3}{c}{Goal} & \multicolumn{3}{c}{Long} \\
\cmidrule(lr){3-5}\cmidrule(lr){6-8}\cmidrule(lr){9-11}\cmidrule(l){12-14}
Model & Attack & SR & \multicolumn{2}{c}{TDS} & SR & \multicolumn{2}{c}{TDS} & SR & \multicolumn{2}{c}{TDS} & SR & \multicolumn{2}{c}{TDS} \\
\midrule
& clean & 97 & $-0.3$ & $_{\pm0.1}$ & 99 & $0.0$ & $_{\pm0.0}$ & 96 & $-0.2$ & $_{\pm0.1}$ & 94 & $-0.1$ & $_{\pm0.1}$ \\
  & noise & 87 & $-0.8$ & $_{\pm0.8}$ & 95 & $-0.2$ & $_{\pm0.2}$ & 81 & $1.1$ & $_{\pm0.6}$ & 52 & $-0.7$ & $_{\pm0.6}$ \\
  & BadWAM & 87 & $-1.0$ & $_{\pm0.6}$ & 97 & $-0.2$ & $_{\pm0.2}$ & 85 & $1.0$ & $_{\pm0.6}$ & 57 & $-1.2$ & $_{\pm0.6}$ \\
\rowcolor{stdred!7}[0pt][\dimexpr2\tabcolsep\relax]\multirow{-4}{*}{FastWAM} & standard & 0 & $87.8$ & $_{\pm0.4}$ & 0 & $71.3$ & $_{\pm0.7}$ & 0 & $90.9$ & $_{\pm0.4}$ & 0 & $58.0$ & \trimrowfill$_{\pm0.8}$ \\
\midrule
& clean & 98 & $0.2$ & $_{\pm0.2}$ & 100 & $0.2$ & $_{\pm0.5}$ & 98 & $0.3$ & $_{\pm0.3}$ & 98 & $0.2$ & $_{\pm0.1}$ \\
  & noise & 86 & $0.5$ & $_{\pm1.1}$ & 98 & $1.0$ & $_{\pm0.3}$ & 92 & $1.4$ & $_{\pm1.3}$ & 75 & $0.3$ & $_{\pm1.0}$ \\
  & BadWAM & 82 & $0.2$ & $_{\pm1.1}$ & 99 & $1.0$ & $_{\pm0.3}$ & 92 & $1.6$ & $_{\pm1.7}$ & 70 & $-0.5$ & $_{\pm1.2}$ \\
\rowcolor{stdred!7}[0pt][\dimexpr2\tabcolsep\relax]& standard & 0 & $73.9$ & $_{\pm0.7}$ & 0 & $63.3$ & $_{\pm0.6}$ & 0 & $76.2$ & $_{\pm0.8}$ & 0 & $55.8$ & \trimrowfill$_{\pm1.4}$ \\
\rowcolor{presblue!7}[0pt][\dimexpr2\tabcolsep\relax]\multirow{-5}{*}{FastWAM-Joint} & preserve & 1 & $62.8$ & $_{\pm1.0}$ & 0 & $38.7$ & $_{\pm0.5}$ & 0 & $67.5$ & $_{\pm1.0}$ & 0 & $47.8$ & \trimrowfill$_{\pm0.6}$ \\
\midrule
& clean & 95 & $0.0$ & $_{\pm0.1}$ & 100 & $0.1$ & $_{\pm0.1}$ & 98 & $-0.1$ & $_{\pm0.1}$ & 98 & $0.0$ & $_{\pm0.1}$ \\
  & noise & 93 & $-1.0$ & $_{\pm0.8}$ & 99 & $-0.3$ & $_{\pm0.3}$ & 82 & $-1.0$ & $_{\pm1.1}$ & 80 & $-0.9$ & $_{\pm0.5}$ \\
  & BadWAM & 92 & $-0.9$ & $_{\pm0.8}$ & 100 & $-0.4$ & $_{\pm0.3}$ & 85 & $-0.9$ & $_{\pm1.1}$ & 80 & $-1.1$ & $_{\pm0.5}$ \\
\rowcolor{stdred!7}[0pt][\dimexpr2\tabcolsep\relax]& standard & 0 & $88.4$ & $_{\pm0.6}$ & 0 & $74.9$ & $_{\pm0.4}$ & 0 & $85.5$ & $_{\pm0.4}$ & 0 & $57.3$ & \trimrowfill$_{\pm1.5}$ \\
\rowcolor{presblue!7}[0pt][\dimexpr2\tabcolsep\relax]\multirow{-5}{*}{Faster-WAM} & preserve & 1 & $80.1$ & $_{\pm0.6}$ & 0 & $64.5$ & $_{\pm0.3}$ & 0 & $84.2$ & $_{\pm0.5}$ & 0 & $59.5$ & \trimrowfill$_{\pm1.0}$ \\
\midrule
& clean & 92 & $0.8$ & $_{\pm0.4}$ & 100 & $-0.2$ & $_{\pm0.3}$ & 98 & $0.1$ & $_{\pm0.4}$ & 98 & $0.3$ & $_{\pm0.2}$ \\
\rowcolor{stdred!7}[0pt][\dimexpr2\tabcolsep\relax]& standard & 2 & $19.9$ & $_{\pm0.4}$ & 6 & $20.3$ & $_{\pm0.4}$ & 11 & $18.2$ & $_{\pm5.2}$ & 4 & $17.8$ & \trimrowfill$_{\pm2.7}$ \\
\rowcolor{presblue!7}[0pt][\dimexpr2\tabcolsep\relax]\multirow{-3}{*}{LingBot-VA} & preserve & 6 & $16.0$ & $_{\pm0.4}$ & 8 & $6.8$ & $_{\pm0.4}$ & 14 & $8.2$ & $_{\pm1.2}$ & 6 & $16.6$ & \trimrowfill$_{\pm2.2}$ \\
\bottomrule
\end{tabular}}
\end{table}

\begin{table}[t]
\caption{\textbf{Attack effectiveness on RoboTwin.} SR is the task success rate (\%) over all tasks. TDS is the commanded end-effector shift along the target axis toward $a^\star$ relative to a paired clean query, in per cent of half the expert range of that axis, and $\pm$ is the error across repeated attack runs. Noise, BadWAM and standard are as in \cref{tab:main}, and preserve uses $\lambda=0.1$. MUT, ASV, VSA and ISO denote the four OpenWAM variants.}
\label{tab:robotwin}
\centering
\small
\setlength{\tabcolsep}{3pt}
\resizebox{\columnwidth}{!}{%
\begin{tabular}{@{}lccccccccc@{}}
\toprule
 & \multicolumn{5}{c}{Reference conditions} & \multicolumn{4}{c}{\textbf{SteerTrace}} \\
\cmidrule(lr){2-6}\cmidrule(l){7-10}
 & Clean & \multicolumn{2}{c}{Noise} & \multicolumn{2}{c}{BadWAM} & \multicolumn{2}{c}{Standard} & \multicolumn{2}{c}{Preserve} \\
\cmidrule(lr){2-2}\cmidrule(lr){3-4}\cmidrule(lr){5-6}\cmidrule(lr){7-8}\cmidrule(l){9-10}
Model & SR & SR & TDS & SR & TDS & SR & TDS & SR & TDS \\
\midrule
FastWAM & 92.0 & 86.5 & 0.0$_{\pm0.0}$ & 84.0 & 0.0$_{\pm0.0}$ & 0.0 & 29.5$_{\pm2.2}$ & 0.0 & 31.1$_{\pm1.9}$ \\
FastWAM-Joint & 94.0 & 94.0 & 0.0$_{\pm0.0}$ & 94.0 & 0.0$_{\pm0.0}$ & 0.0 & 29.3$_{\pm1.0}$ & 4.5 & 15.6$_{\pm0.8}$ \\
Faster-WAM & 91.5 & 90.5 & 0.0$_{\pm0.1}$ & 89.5 & 0.0$_{\pm0.1}$ & 0.5 & 31.5$_{\pm1.3}$ & 0.0 & 25.3$_{\pm1.2}$ \\
\addlinespace[2pt]
OpenWAM-MUT & 91.0 & 91.5 & 0.2$_{\pm0.1}$ & 94.0 & 0.1$_{\pm0.1}$ & 1.0 & 51.0$_{\pm1.9}$ & 6.0 & 31.3$_{\pm2.3}$ \\
OpenWAM-ASV & 93.5 & 93.5 & 0.1$_{\pm0.1}$ & 92.0 & 0.0$_{\pm0.1}$ & 2.0 & 62.9$_{\pm2.1}$ & 1.5 & 63.5$_{\pm1.9}$ \\
OpenWAM-VSA & 89.0 & 85.5 & 0.1$_{\pm0.0}$ & 88.5 & 0.1$_{\pm0.1}$ & 0.5 & 85.0$_{\pm2.9}$ & 0.0 & 81.1$_{\pm3.0}$ \\
OpenWAM-ISO & 87.0 & 86.5 & 0.1$_{\pm0.1}$ & 88.0 & 0.2$_{\pm0.1}$ & 0.0 & 76.8$_{\pm3.0}$ & 0.5 & 79.5$_{\pm3.2}$ \\
\bottomrule
\end{tabular}}
\end{table}

\nbf{Task Disruption.}
For each model, a single universal perturbation, optimized on that model and shared across all tasks, makes it fail. On LIBERO, the standard attack lowers the success rate from $94$--$100\%$ to $0\%$ on all four suites for FastWAM, FastWAM-Joint and Faster-WAM, and to $2.0\%$ and $6.5\%$ on Spatial and Object for LingBot-VA (\cref{tab:main}). On RoboTwin, it lowers the success rate of all six models from $87.0$--$93.5\%$ to $0$--$2\%$ (\cref{tab:robotwin}). The same-budget baselines are far weaker. Gaussian noise and BadWAM~\cite{badwam} leave the success rate at $52$--$100\%$ on LIBERO and $84$--$94\%$ on RoboTwin.

\nbf{Targeted Action Control.}
Our attack also steers the executed action toward $a^\star$. Its TDS is $55.8$--$90.9$ for the three LIBERO models and $19.9$--$20.3$ for LingBot-VA (\cref{tab:main}), and $29.5$--$85.0$ on RoboTwin (\cref{tab:robotwin}). The baselines move the attacked action by at most $1.6$ on LIBERO and $0.2$ on RoboTwin, in either direction, even where they lower the success rate. Task failure and targeted control are therefore distinct: the baselines cause some of the former, but only our attack produces the latter. The following analyses trace this targeted effect through the model.

\subsection{Preserving the predicted future}
\label{sec:preserve}

\begin{figure*}[t]
\centering
\includegraphics[width=\textwidth]{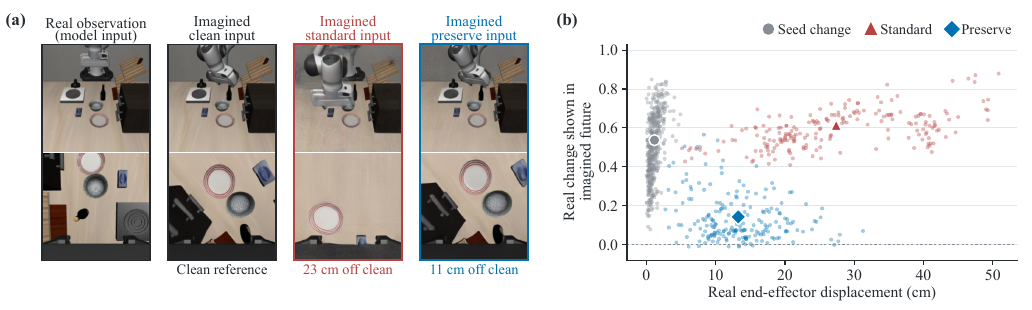}
\caption{\textbf{The imagined future does not show the executed action (FastWAM-Joint, LIBERO).} (a)~Model input and last imagined frame for clean, standard and preserve inputs (third-person view top, wrist view bottom), with the end-effector distance from clean execution below. (b)~One point per executed chunk from $200$ initial states: end-effector displacement from clean execution versus the projection ratio of imagined onto real visual change. Large markers are group means.}
\label{fig:imagsim}
\end{figure*}

The preserve attack adds the future-preserving loss of \cref{eq:preserve} to the standard objective. 

\begin{table}[t]
\caption{\textbf{Predicted future fidelity.} Lower is better. Noise is same-budget Gaussian noise. Std.\ and Pres.\ denote the standard and preserve attacks. Full results are in Supplementary.}
\label{tab:future_fidelity}
\centering
\small
\setlength{\tabcolsep}{2pt}
\begin{tabular}{@{}lcccccc@{}}
\toprule
 & \multicolumn{3}{c}{$\Dfut\downarrow$} & \multicolumn{3}{c}{LPIPS$\downarrow$} \\
\cmidrule(lr){2-4}\cmidrule(l){5-7}
Model & Noise & Std. & Pres. & Noise & Std. & Pres. \\
\midrule
\multicolumn{7}{@{}l}{\emph{LIBERO}} \\
FastWAM-Joint & 0.65 & 0.87 & 0.53 & 0.187 & 0.329 & 0.131 \\
Faster-WAM & 0.60 & 0.88 & 0.55 & 0.174 & 0.365 & 0.148 \\
LingBot-VA & 0.40 & 0.64 & 0.46 & 0.053 & 0.122 & 0.079 \\
\midrule
\multicolumn{7}{@{}l}{\emph{RoboTwin}} \\
FastWAM-Joint & 0.31 & 0.66 & 0.50 & 0.059 & 0.383 & 0.234 \\
Faster-WAM & 0.30 & 0.70 & 0.56 & 0.054 & 0.526 & 0.336 \\
OpenWAM-MUT & 0.34 & 0.67 & 0.36 & 0.089 & 0.473 & 0.136 \\
OpenWAM-ASV & 0.35 & 0.71 & 0.57 & 0.091 & 0.565 & 0.438 \\
OpenWAM-VSA & 0.36 & 0.62 & 0.44 & 0.098 & 0.356 & 0.180 \\
OpenWAM-ISO & 0.35 & 0.62 & 0.42 & 0.091 & 0.370 & 0.149 \\
\bottomrule
\end{tabular}
\end{table}

\nbf{The predicted future returns toward clean.}
The preserve term improves both fidelity metrics over the standard attack in every model (\cref{tab:future_fidelity}). On LIBERO, FastWAM-Joint and Faster-WAM obtain a future closer to clean than same-budget noise on both metrics: $\Dfut$ falls from $0.87$ and $0.88$ to $0.53$ and $0.55$, against $0.65$ and $0.60$ for noise. On FastWAM-Joint, the preserve future is even closer to clean than that of BadWAM on both metrics ($\Dfut=0.53$ vs.\ $0.57$), BadWAM produces little action shift in the target direction($|\mathrm{TDS}|\le1.6$) while the preserve attack shifts it by $38.7$--$67.5$ (\cref{tab:main}). LingBot-VA also improves ($\Dfut$ from $0.64$ to $0.46$) but stays above noise ($0.40$). On RoboTwin, the preserve term lowers $\Dfut$ by $20$--$46\%$, yet every model remains above the noise level. OpenWAM \texttt{mutual} comes closest ($\Dfut=0.36$ vs.\ $0.34$), and \texttt{action\_sees\_video} recovers least ($\Dfut=0.57$).

\nbf{The executed action is still steered.}
On LIBERO, the preserve attack keeps the success rate at $0$--$1\%$ on every suite for FastWAM-Joint and Faster-WAM, and retains $61$--$89\%$ and $86$--$104\%$ of the standard TDS on each suite (\cref{tab:main}). LingBot-VA retains $80\%$ on Spatial at a success rate of $6\%$. On RoboTwin, \texttt{video\_sees\_action} and \texttt{isolated} reach success rates of $0\%$ and $0.5\%$, respectively. 

\nbf{The imagined future no longer shows what the robot does.}
Executing the predicted actions in the simulator makes this gap concrete (\cref{fig:imagsim}). On $200$ initial states of LIBERO, we predict and execute the action chunk of FastWAM-Joint under the clean, standard and preserve conditions, and compare the imagined future with the frames recorded during execution by the pixel RMSE of the full RGB frames at execution steps $4,8,\ldots,32$. We also measure the share of the real change, relative to the clean execution, that the imagined future shows. Under the standard attack, the imagined future is closer to the execution of the attacked action than to that of the clean action in $186$ states ($93\%$), and it shows $61\%$ of the real change, about as much as a change of sampling seed ($54\%$). Under the preserve attack, the imagined future is instead closer to the clean execution in $186$ states ($93\%$) and shows only $14\%$ of the real change, while the end effector still ends $13.3$\,cm from the clean execution on average ($27.4$\,cm under the standard attack). A monitor that inspects only the imagined future may thus see a nearly normal future while the robot acts differently.

\subsection{Tokens that carry the targeted effect}
\label{sec:route}

We now ask which token groups carry the targeted effect and whether the preserve objective changes them. For the current-frame tokens $\mathcal{C}$ and the future tokens $\mathcal{F}$, we restore their clean K/V ($\Nec_G$) or inject their attacked K/V into a clean run ($\Suf_G$, \cref{eq:nec,eq:suf}).

\begin{table}[t]
\caption{\textbf{Pathways of successful attacks.} LIBERO for FastWAM-Joint, Faster-WAM and LingBot-VA, RoboTwin for OpenWAM. $\Nec_{\mathcal{F}}$ and $\Nec_{\mathcal{C}}$ are the percentages of the targeted effect removed by restoring the future or current-frame K/V to clean. $\Suf_{\mathcal{F}}$ is the percentage reproduced by injecting only the attacked future K/V into a clean pass. $^\ast$For LingBot-VA, $\Nec_{\mathcal{C}}$ restores the most recent observed video tokens in the history cache.}
\label{tab:path}
\centering
\small
\setlength{\tabcolsep}{2.5pt}
\begin{tabular}{@{}lcccccc@{}}
\toprule
 & \multicolumn{2}{c}{Restore future} & \multicolumn{2}{c}{Restore current} & \multicolumn{2}{c}{Inject future} \\
\cmidrule(lr){2-3}\cmidrule(lr){4-5}\cmidrule(l){6-7}
Model & Std. & Pres. & Std. & Pres. & Std. & Pres. \\
\midrule
FastWAM-Joint & \textbf{64} & 12 & 47 & \textbf{99} & 53 & 1 \\
Faster-WAM & \textbf{81} & 15 & 39 & \textbf{87} & 61 & 13 \\
OpenWAM-MUT & 61 & 9 & \textbf{76} & \textbf{95} & 24 & 5 \\
OpenWAM-ASV & \textbf{94} & \textbf{93} & 10 & 9 & 90 & 91 \\
LingBot-VA$^\ast$ & \textbf{83} & \textbf{76} & 2 & 2 & 78 & 69 \\
\bottomrule
\end{tabular}
\end{table}

\nbf{Unconstrained attacks pass through the imagined future.}
In all five models whose action expert attends to the imagined future, restoring the clean future K/V removes $61$--$94\%$ of the targeted effect of the standard attack (\cref{tab:path}). The future is therefore not only displaced by the attack (\cref{tab:future_fidelity}) but carries most of the targeted effect to the action. Models without this attention are analyzed in the Supplementary.

\nbf{Preserving the future moves the attack to the current frame.}
On FastWAM-Joint, Faster-WAM and OpenWAM-MUT, restoring the future K/V removes only $9$--$15\%$ of the targeted effect of the preserve attack, whereas restoring the current-frame K/V removes $87$--$99\%$, up from $39$--$76\%$ under the standard attack. In absolute terms, the future pathway nearly vanishes: on FastWAM-Joint, restoring the future removes $0.048$ of $T$ instead of $0.436$, while restoring the current frame removes $0.404$ instead of $0.320$. On OpenWAM-MUT, restoring the future removes $0.018$ instead of $0.221$. OpenWAM-ASV and LingBot-VA keep passing through the future (\cref{tab:path}).

\begin{figure}[t]
\centering
\includegraphics[width=\columnwidth]{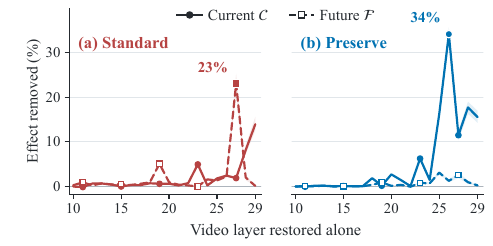}
\caption{\textbf{Preserve moves the targeted effect from $\mathcal{F}$ to $\mathcal{C}$.} Percentage of the targeted effect removed in FastWAM-Joint when the action expert reads clean keys and values at a single video layer, for the current-frame tokens $\mathcal{C}$ (solid) or the future tokens $\mathcal{F}$ (dashed), under the standard (red) and preserve (blue) attacks. The standard attack peaks on $\mathcal{F}$ at layer 27 ($23\%$), and the preserve attack peaks on $\mathcal{C}$ at layer 26 ($34\%$). Restoring layers 25--29 together removes $34\%$ of the targeted effect of the standard attack through $\mathcal{C}$ and $42\%$ through $\mathcal{F}$, against $82\%$ and $8\%$ of that of the preserve attack. Layers 0--9 remove less than $1.5\%$ in every condition and are omitted.}
\label{fig:layerwise_kv}
\end{figure}

\nbf{The attack switches token group at the same layers.}
Restoring one layer at a time in FastWAM-Joint shows where each pathway enters the action expert (\cref{fig:layerwise_kv}). Under the standard attack, the largest single-layer effect lies on the future tokens at layer 27 ($23\%$), and the current-frame tokens matter only in the last two layers ($8\%$ and $14\%$). Under the preserve attack, no layer of the future tokens removes more than $3\%$, while the current-frame tokens carry the targeted effect in layers 25--29, peaking at layer 26 ($34\%$). Restoring these five layers of the current frame together removes $82\%$ of the targeted effect of the preserve attack, against $34\%$ under the standard attack. The attack thus keeps entering the action expert in the last layers but switches from the future tokens to the current-frame tokens. The same five layers of the current frame remove $75\%$ of the targeted effect of the preserve attack on Faster-WAM ($23\%$ under standard) and $85\%$ on OpenWAM-MUT ($73\%$).

\subsection{Neurons that carry the targeted effect}
\label{sec:neurons}

Having located the token pathways, we ask whether the targeted effect depends on a small set of FFN neurons in the video expert, and whether these are the neurons whose activations change most.

\begin{figure}[t]
\centering
\includegraphics[width=\columnwidth]{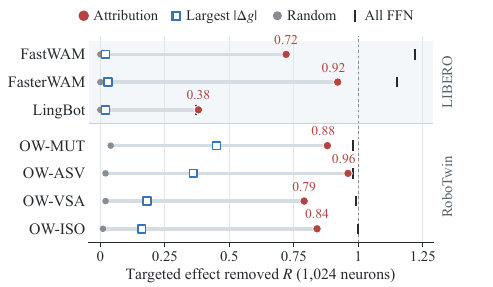}
\caption{\textbf{Attributed neurons carry the targeted effect.} Fraction of the targeted effect removed, $R$ (\cref{eq:R}), under the standard attack when $1{,}024$ FFN neurons ($0.36\%$) of video layers 10--29 are restored to clean, selected by attribution (\cref{eq:attr}), by the largest activation change $|\Delta g|$, or at random. ``All FFN'' restores every neuron in these layers. OW denotes OpenWAM. Results for other $k$ are in the Supplementary.}
\label{fig:neurons}
\end{figure}

\nbf{A small set of neurons carries the targeted effect in every task.}
We select the $1{,}024$ highest-attribution neurons once over all tasks, $0.36\%$ of the FFN neurons in layers 10--29. Restoring them removes $72$--$96\%$ of the targeted effect in six models, and in LingBot-VA as much as all FFN neurons in these layers (\cref{fig:neurons}). Equally many random neurons remove at most $4\%$, and the neurons with the largest activation change remove far less than the attributed ones in every model. Attribution also agrees in rank with the measured effect of single neurons, with details in the Supplementary. The same set works on each task: it removes at least $70\%$ of the targeted effect on each of the 50 RoboTwin tasks in all four OpenWAM variants. Under both attacks, selecting the neurons without the evaluated task removes as much of its targeted effect as selecting them over all tasks. A universal perturbation thus steers every task through the same small set of neurons.

\nbf{The neuron set follows the pathway.}
When the attack changes pathway, it also changes neurons. On FastWAM-Joint and Faster-WAM, where the preserve attack leaves the future (\cref{sec:route}), the set of the other attack removes only $16$--$35\%$ of the targeted effect, against $91$--$107\%$ for its own set. On OpenWAM-ASV, OpenWAM-VSA and OpenWAM-ISO, whose pathway does not change, the two sets are interchangeable ($70$--$95\%$).

\nbf{Token-specific neurons.}
Restricting neuron restoration to the current-frame or the future tokens shows the same switch inside the video expert. On FastWAM-Joint, restoring the top $1{,}024$ neurons on the future tokens removes more of the targeted effect than on the current-frame tokens under the standard attack ($80.6\%$ vs.\ $37.6\%$), and the order reverses under the preserve attack ($25.3\%$ vs.\ $82.5\%$). On Faster-WAM and OpenWAM-MUT, the current-frame tokens already dominate under the standard attack, and under the preserve attack the future tokens remove almost nothing ($32\%\to2\%$ and $38\%\to3\%$). Full results for other $k$ and for injection are in the Supplementary.

\subsection{Directions that carry the targeted effect}
\label{sec:directions}

The perturbation also changes the residual stream of every video layer (the residual shift, \cref{sec:mech}). We ask whether the directions that hold most of its energy also carry the targeted effect on the action.

\begin{table}[t]
\caption{\textbf{Directions 2--16 carry the targeted effect.} Energy share $\rho$, necessity $\mathcal{N}$ and sufficiency $\Sigma$ (\%, \cref{sec:mech}) of bands of residual-shift eigendirections under the standard attack. Random: 15 random directions removed instead of $u_{2\text{--}16}$. LIBERO for FastWAM, FastWAM-Joint, Faster-WAM and LingBot-VA, RoboTwin for OpenWAM.}
\label{tab:subspace}
\centering
\small
\setlength{\tabcolsep}{4pt}
\begin{tabular}{@{}lcccccc@{}}
\toprule
 & \multicolumn{2}{c}{Energy $\rho$} & \multicolumn{2}{c}{Remove, $\mathcal{N}$} & \multicolumn{2}{c}{Keep only, $\Sigma$} \\
\cmidrule(lr){2-3}\cmidrule(lr){4-5}\cmidrule(l){6-7}
Model & $u_1$ & $u_{2\text{--}16}$ & $u_{2\text{--}16}$ & Rand. & $u_1$ & $u_{1\text{--}64}$ \\
\midrule
FastWAM & 66 & 14 & 96 & 16 & 0 & 18 \\
FastWAM-Joint & 64 & 11 & 97 & 13 & 0 & 13 \\
Faster-WAM & 45 & 16 & 97 & 13 & 0 & 13 \\
LingBot-VA & 41 & 14 & 73 & 9 & 7 & 84 \\
OpenWAM-MUT & 44 & 18 & 99 & 39 & 0 & 4 \\
OpenWAM-ASV & 48 & 17 & 98 & 24 & 0 & 11 \\
OpenWAM-VSA & 44 & 16 & 99 & 15 & 0 & 6 \\
OpenWAM-ISO & 44 & 16 & 98 & 14 & 0 & 6 \\
\bottomrule
\end{tabular}
\end{table}

\nbf{Low-energy directions carry the targeted effect.}
Where the perturbation changes the residual stream most is not where it steers the action. Across models, the leading direction $u_1$ holds $41$--$66\%$ of the residual shift energy, but keeping it alone reproduces at most $7\%$ of the targeted effect (\cref{tab:subspace}). Directions 2--16, which hold only $11$--$18\%$ of the energy, are instead necessary: removing them cancels $96$--$99\%$ of the targeted effect in seven models, against $13$--$39\%$ for 15 random directions, and the same band is found when the evaluated task is left out. The overall magnitude of the residual shift likewise does not indicate whether a perturbation steers the action, with details in the Supplementary.

\section{Conclusion}
SteerTrace steers eight WAMs toward an attacker-chosen action, and its signed effect lets us trace how it reaches the action. It passes through the imagined future when it can and moves to the current frame when the future is kept clean. A set of $0.36\%$ of the FFN neurons and a 15-dimensional low-energy residual subspace carry most of the targeted effect. In closed-loop evaluation, a clean-looking imagined future can hide substantial action shifts.

{\small
\bibliographystyle{ieeenat_fullname}
\bibliography{refs}
}

\end{document}